\documentclass[a4paper, 10 pt, conference]{ieeeconf}  

\IEEEoverridecommandlockouts                              

\usepackage{cite}
\usepackage{amsmath,amssymb,amsfonts}
\usepackage{algorithmic, amssymb, gensymb}
\usepackage{graphicx}
\usepackage{textcomp}
\usepackage{xcolor}
\def\BibTeX{{\rm B\kern-.05em{\sc i\kern-.025em b}\kern-.08em
    T\kern-.1667em\lower.7ex\hbox{E}\kern-.125emX}}

\usepackage[breaklinks=true,bookmarks=false]{hyperref}
\usepackage{tikz,pgfplots}
\pgfplotsset{compat=1.18}
\usepackage{multirow}

\usepackage{subcaption}

\title{\LARGE \bf
Good Grasps Only: A data engine for self-supervised fine-tuning of pose estimation using grasp poses for verification
}

\author{Frederik Hagelskjær 
\thanks{This project was funded in part by Innovation Fund Denmark through the projects MADE FAST and FERA, and in part by the SDU I4.0-Lab.}
\thanks{
All authors are with SDU Robotics, Mærsk Mc-Kinney Møller Institute, University of Southern Denmark, 5230 Odense M, Denmark
    {\tt frhag@mmmi.sdu.dk}}%
}

\begin{document}

\maketitle
\thispagestyle{empty}
\pagestyle{empty}


\begin{abstract}
In this paper, we present a novel method for self-supervised fine-tuning of pose estimation. Leveraging zero-shot pose estimation, our approach enables the robot to automatically obtain training data without manual labeling. After pose estimation the object is grasped, and in-hand pose estimation is used for data validation. Our pipeline allows the system to fine-tune while the process is running, removing the need for a learning phase.

The motivation behind our work lies in the need for rapid setup of pose estimation solutions. Specifically, we address the challenging task of bin picking, which plays a pivotal role in flexible robotic setups. 

Our method is implemented on a robotics work-cell, and tested with four different objects. For all objects, our method increases the performance and outperforms a state-of-the-art method trained on the CAD model of the objects. Project page available at \href{https://gogoengine.github.io}{gogoengine.github.io}











\end{abstract}



    

    

\section{Introduction}
The automation of industrial processes enables much greater output of production at reduced prices. 
Successful automation is most often obtained for large batch productions, as the large production output allows for the huge development cost. 
But, small batch productions constitutes a large part of production tasks. And for small batches the huge development cost cannot be tolerated. Flexible set-ups using, e.g. robotics for shorter set-up times, are thus important \cite{yokokohji2022world}. 

One important aspect of flexible set-ups is the feeding process \cite{yokokohji2019assembly}. To allow for fully automatic processes, no manual insertion of objects should be required. While mechanical solutions such as bowl feeders have been created, they all require manual labor for configuration \cite{mathiesen2018optimisation}.

One solution that does not require any physical configuration is visual pose estimation for bin-picking. However, visual pose estimation often requires a lot of parameter tuning to obtain usable performance \cite{hagelskjaer2018does}. This added set-up time can make the solution unfeasible. 


One solution to avoid manual tuning is using deep learning to obtain the parameters. However, manually collecting data is generally time-consuming and limits the usability of solutions. To overcome this, synthetic data generation has been introduced, however, a reality gap exists between the real and synthetic data. This reality gap often decreases the performance of the algorithm \cite{hodavn2020bop, wang2021occlusion}. These methods also require simulation of the data, and a subsequent training phase, both of which increase the set-up time.

\begin{figure}[t]
    \vspace{2mm}
    \begin{center}
       \includegraphics[width=0.99\linewidth]{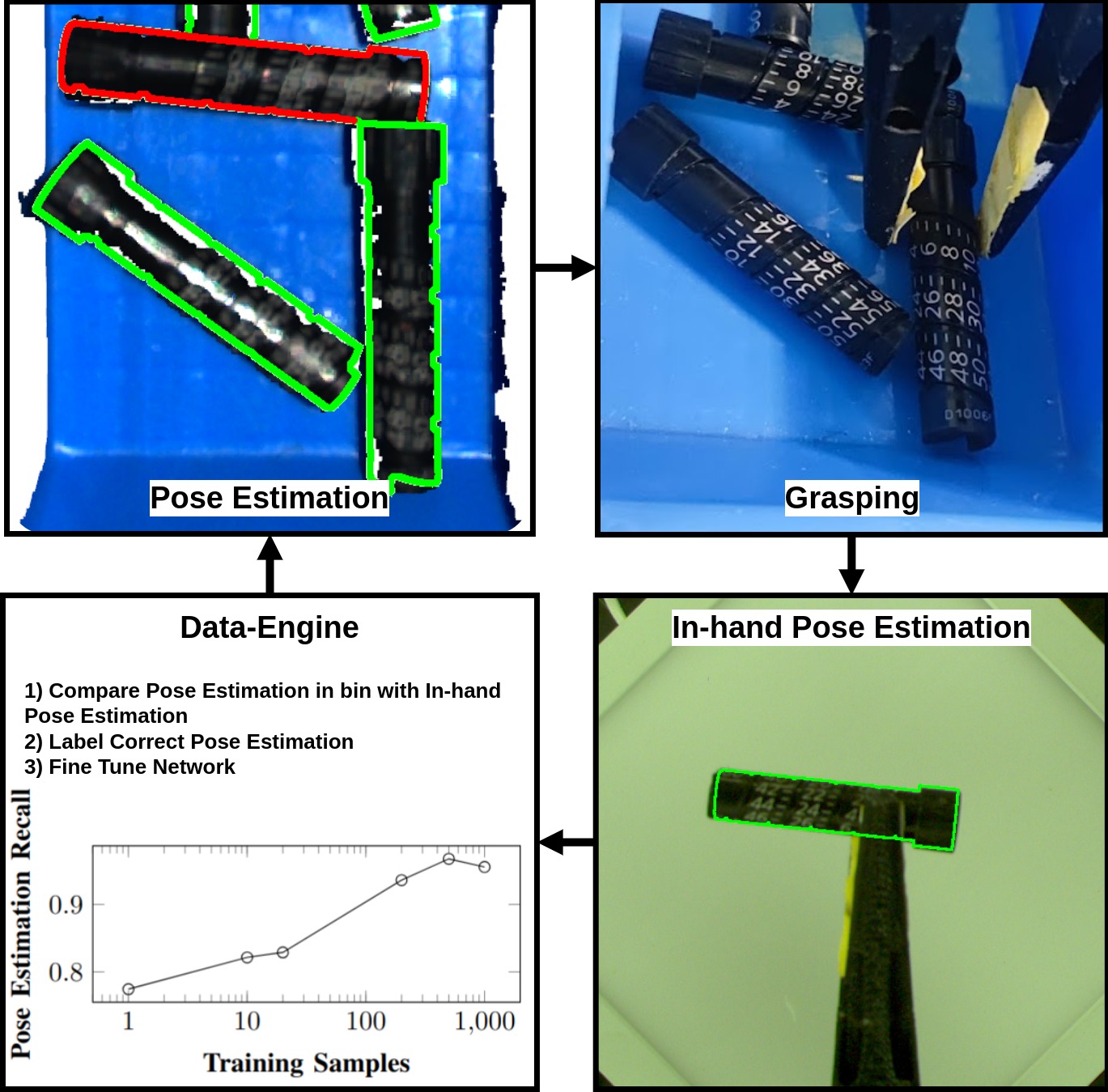}
       \caption{The pipeline of our developed data engine. First using zero-shot a pose estimation is performed. Then the object is grasped and an in-hand pose estimation is performed. Comparing the two poses, correct pose estimations are sorted, and the network is fine-tuned. A new pose estimation is then performed and the process repeats. As data is gradually collected the network performance increases.}
       \label{fig:pecomp}
        \vspace{-6mm}
    \end{center}
\end{figure}




Zero-shot methods, which does not require training for new objects, have shown promising results in this regard \cite{hodan2024bop}. However, methods trained with data of the actual object generally obtain better performance \cite{hodan2024bop, wang2020self6d}.
In this regard self-supervised methods have shown very good results in obtaining training data automatically \cite{deng2020self, mitash2017self, wang2021occlusion, yu2023robotic}. However, the data collection is generally performed in a training phase where the robot is not operable \cite{deng2020self}. This training phase is necessary as the data cannot be collected while the system is running. This is either because the system cannot gather data while executing the task, or because the system will not work with the initial performance. As incorrect pose estimations could result in non-recoverable errors. As the system's feasibility depends on the set-up time, this learning phase limits the usability of such systems. 





In this paper, a method for online self-supervised learning of pose estimation for bin-picking is presented. The self-supervised method is built into the task, thus allowing the system to start the task immediately and then gradually improve performance. The methodology is visualized in Fig.~\ref{fig:pecomp}.

Our method uses an existing workcell for bin-picking \cite{hagelskjaer2024off} combined with an existing in-hand pose estimation system \cite{hagelskjaer2022hand}. Additionally, we replace the network of the bin-picking pose estimation with a zero-shot method \cite{hagelskjaer2023keymatchnet}. 

The presented data engine can work with any combination of pose estimation based bin-picking and in-hand verification, however, the presented system allows several advantages.
The bin picking system is able to recover from failed grasps \cite{hagelskjaer2024off}, thus allowing the system to run with imperfect pose estimations. This along with using a zero-shot pose estimation method allows the system to run without any configuration. 
%
And as the method does not use color information, simple industrial CAD models can be used.

After the object has been grasped, an in-hand pose estimation verifies the object pose. The in-hand pose estimation combines template matching with stable poses. The template matching does not require any configuration or training, while the stable poses makes the solution very robust \cite{hagelskjaer2019using, hagelskjaer2022hand}. This makes the set-up of the in-hand pose estimation very simple.

An additional strategy to improve the abilities of the system when starting out, is to set the initial pose estimation parameters to prioritize precision and disregard the run-time \cite{hagelskjaer2022parapose}. While a long cycle-time can impact the overall usability of the system, the long cycle-time is only used in the beginning. As data is collected the performance of the pose estimation improves and the cycle-time can gradually be reduced.
These parts combined makes the system able to run immediately without tuning the system. We demonstrate that the method is able to work for four different objects, and that the performance is able to increase as the system runs. We also show that the system is able to learn generalities from the scene and improve performance for unseen objects.

%







The main contributions presented in this paper are:
\begin{itemize}
    \item A method for creating ground-truth data for bin-picking
    \item A pose estimation data engine which allow for self-supervised learning during task execution 
    \item Integration of the data engine on a work-cell for bin-picking and in-hand pose estimation
    \item Experiments demonstrating the effectiveness of the method
\end{itemize}

The remaining paper is structured as follows: related papers are reviewed in Sec.~\ref{related}. In Sec.~\ref{met}, the work-cell and our developed method are elaborated. In Sec.~\ref{exp}, experiments are performed, showing the validity of the approach. Finally, in Sec.~\ref{con}, a conclusion is given, and further work is discussed.

\section{Related Work}
\label{related}
Bin-picking is an important tool for flexible robotic systems \cite{yokokohji2019assembly}. As the set-up of bin-picking is generally very time-consuming, many different approaches for automatic set-ups have been developed.
%
%

One approach is to avoid the pose estimation altogether, using model-free bin-picking \cite{cordeiro2022bin}. To improve the performance of such methods many different approaches to self-supervised bin-picking have been created \cite{berscheid2019improving, suzuki2020online, shao2019suction}.
%
An example is shown in \cite{berscheid2020self} where one-shot imitation learning is combined with self-supervised learning. Using a demonstrated goal state the robot automatically learns to grasp the objects correctly. Another approach is shown in \cite{zhao2022learning}, where the learning is performed using simulated data. These methods show very good results and the approach seems very promising.
However, model-free bin-picking is not a feasible solution to our scenario. As the manipulation of the objects can only be performed from a limited number of grasp poses the grasping needs to be guided by the object's position.
%


To obtain precise pose estimates for the data engine, we use KeyMatchNet \cite{hagelskjaer2023keymatchnet}, a colorless zero-shot algorithm. While other zero-shot pose estimation methods exist, they require color information and are generally not independent of detectors \cite{hodan2024bop}. Both are requirements for our data engine. For a more in-depth explanation of zero-shot methods see \cite{hodan2024bop, hagelskjaer2023keymatchnet}.

\subsection{Self-supervised Pose Estimation}
%
Self-supervised learning of pose estimation allows for increased performance without any manual labor. As a result, many different applications have been developed.
One such method to automatically collect data for self-supervised learning of pose estimation is presented in \cite{deng2020self}. The network is first trained using synthetic data. 
Similar to our method, this network is then used to obtain the pose estimations for fine-tuning the network. Dissimilar to our method, they use object tracking to create the dataset, where we instead reject false positives. 
As the initial network cannot handle difficult cases, curriculum learning is employed to start the process with simple scenes of a single object. The complexity of the scenes is then increased as the data is collected. 
Our method does not require this learning phase and instead starts with the actual task. As our system can automatically remove false positives we robust towards wrong pose estimations.


The method presented in \cite{mitash2017self} is also initially trained on synthetic data, but also adds physical simulation of the object positions. Similar to our method they employ a more expensive pose estimation system for data collection, however, they use multi-view for verification as opposed to our in-hand pose estimation. 
Similar to \cite{deng2020self} the data is collected in a learning phase, and not when performing the task.

Another method pre-trained on synthetic data is presented in \cite{wang2020self6d}. By using neural rendering with real, but unlabeled, images, they improve the pose estimation performance. The method is tested using a benchmarking dataset instead of a robotic set-up, and obtains good performance. However, the performance does not equate to methods trained on the ground truth information from the real data.


\begin{figure*}[t]
    \vspace{1.5mm}
    \begin{center}
    \hfill
    \begin{subfigure}[t]{.33\textwidth}
      \centering
      \includegraphics[trim=0 0 0 475,clip, width=0.99\linewidth]{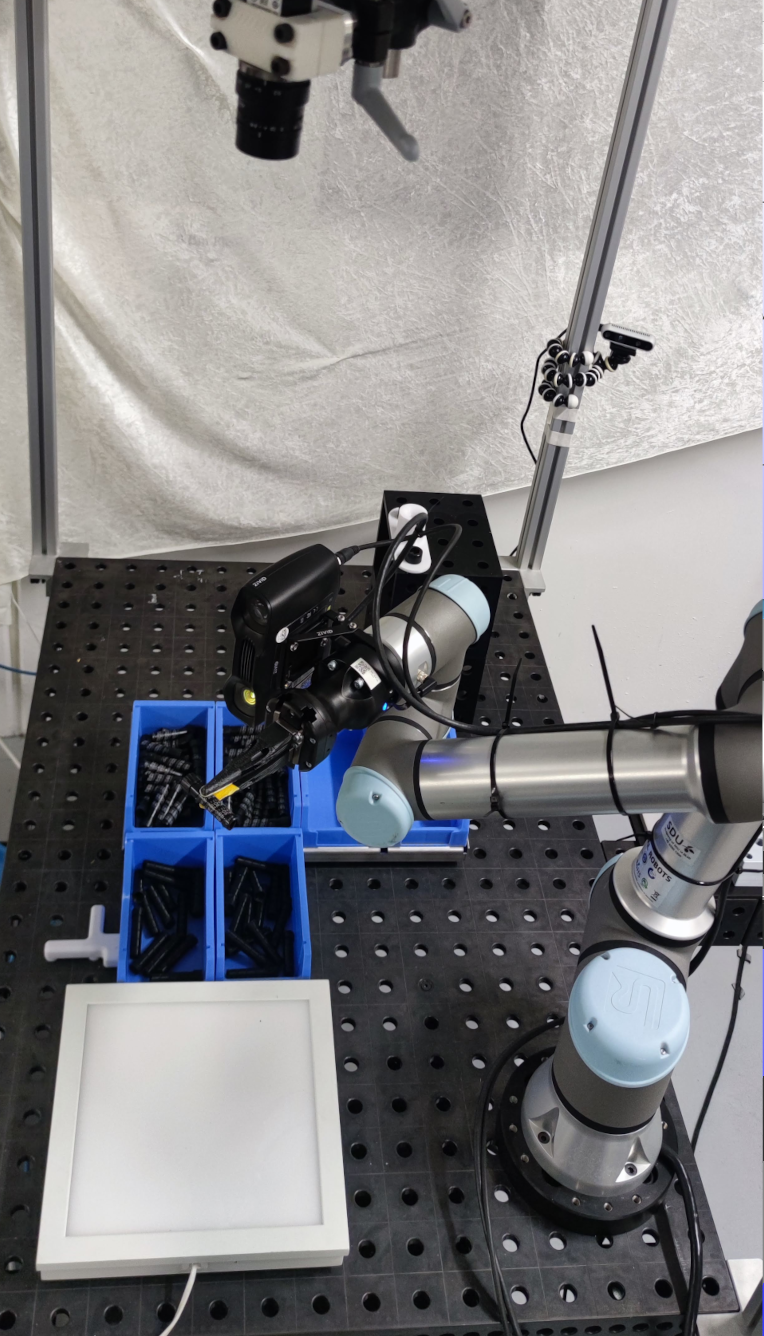}
      \caption{Real workcell.}
      \label{fig:workcell:01}
    \end{subfigure}%
    \hfill
    \begin{subfigure}[t]{.33\textwidth}
      \centering
      \includegraphics[trim=0 0 0 475,clip, width=0.99\linewidth]{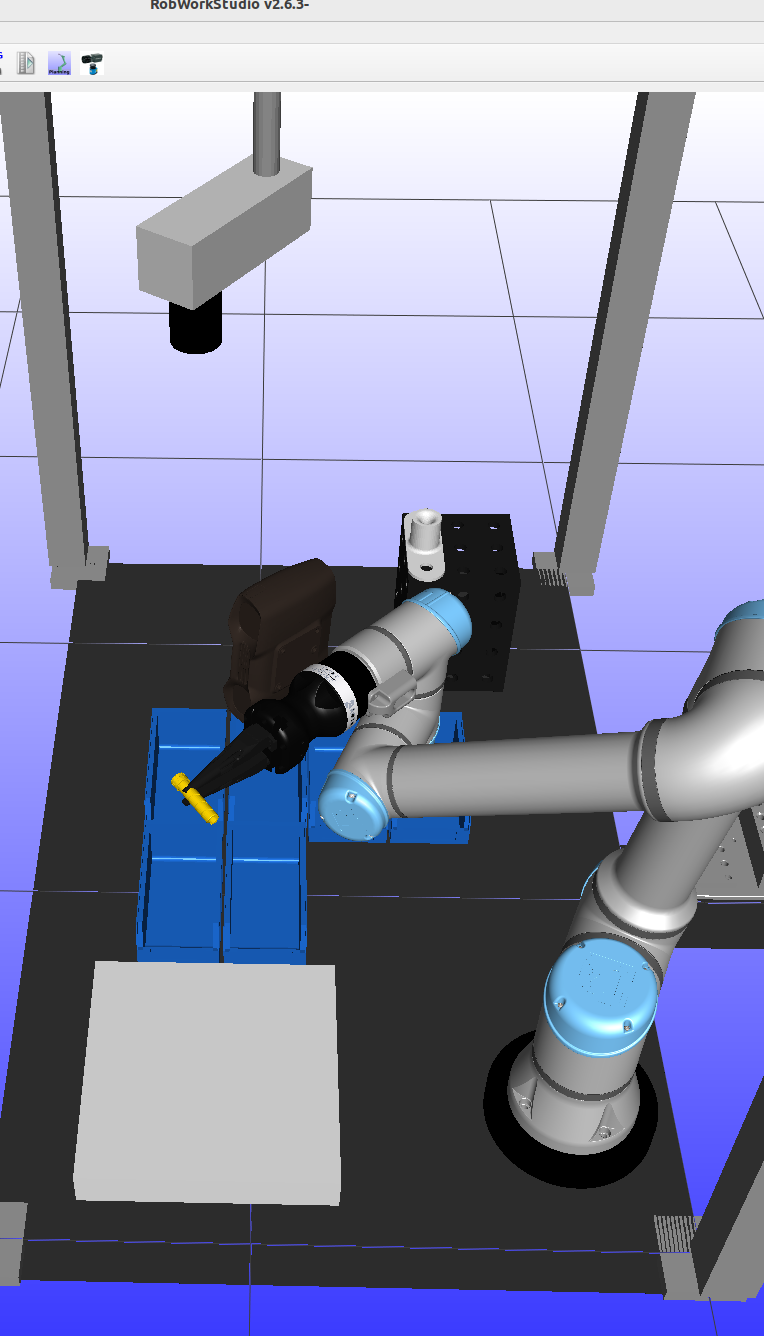}
      \caption{Digital twin.}
      \label{fig:workcell:02}
    \end{subfigure}%
    \hfill
    \begin{subfigure}[t]{.264\textwidth}
      \centering
      \includegraphics[trim=70 0 270 0,clip, width=0.99\linewidth]{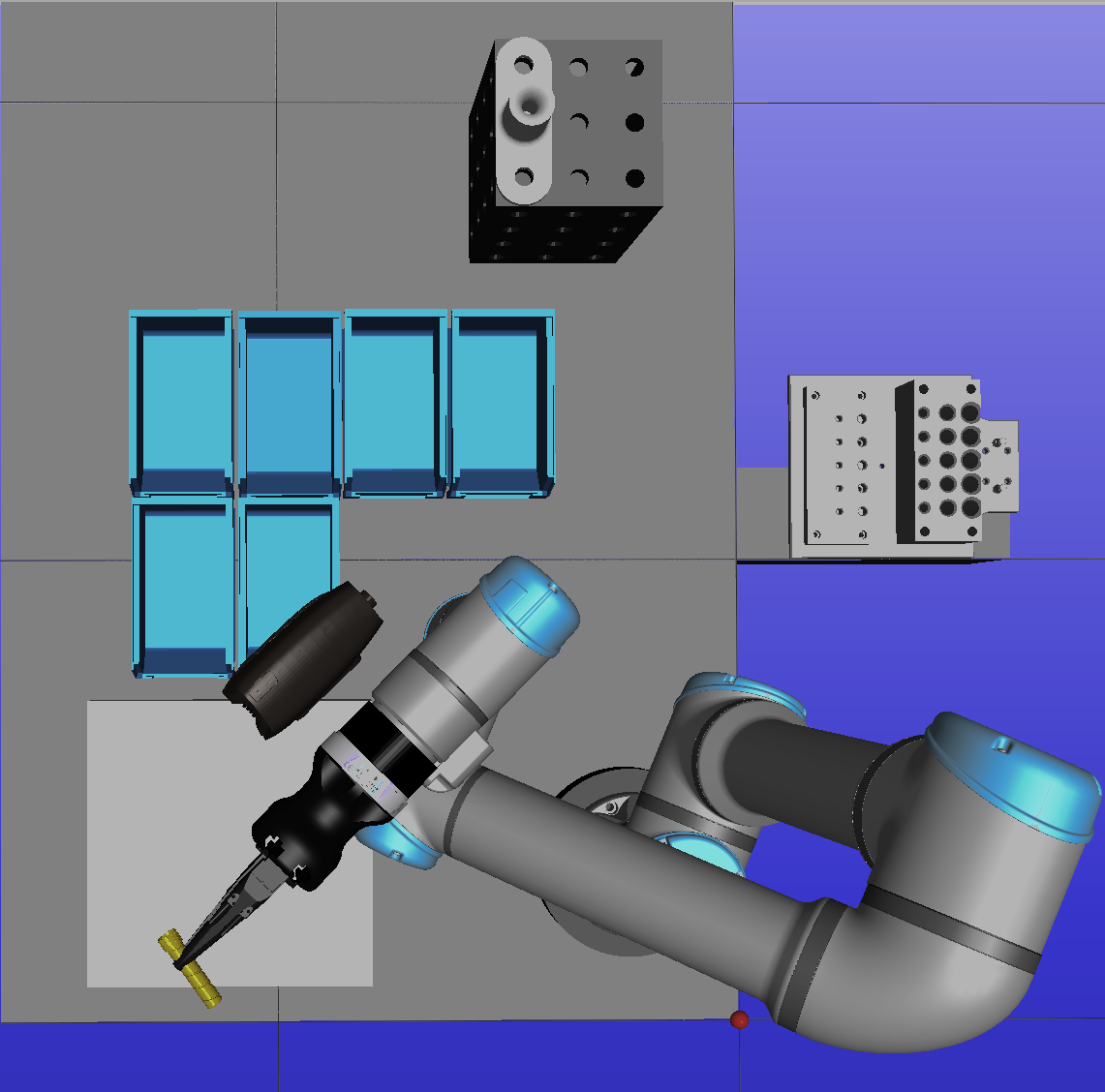}
      \caption{Digital twin from above.}
      \label{fig:workcell:03}
    \end{subfigure}%
    \hfill
    ~

   \caption{The workcell used for experiments. The real workcell is shown in Fig.~\ref{fig:workcell:01} with a digital twin shown in Fig.~\ref{fig:workcell:02}. The digital twin allows for planning collision-free movements, as robot the movements are dependent on the found object poses. A top view is shown in Fig.~\ref{fig:workcell:03}. The object bins are placed in the center. At the bottom left the background light for the in-hand pose estimation is located. The fixture is shown at the top right.
   }
   \label{fig:workcell}
    \vspace{-6mm}
\end{center}
\end{figure*}

Similar to our method SCNet \cite{yu2023robotic}, use a PointNet \cite{qi2017pointnet} like structure for the pose estimation. However, the focus is on category-level pose estimation, and they disregard the CAD model. However, as the CAD model is available, we utilize this data. 



\subsection{Data-Engine}
The automatic collection of data is an integral part of our method. Compared with to our method, many different approaches for data collection and learning have been developed. 
One solution is human-robot interaction \cite{boschetti2023improving}. Here a human assists the robot when any failures occur. The feedback from the human is not only used to solve the occurring failure but also to learn and improve future performance. This approach could potentially be used for our system, but as the set-up should be fully automatic we have not employed any manual correction.

A different method to automatically obtain data is by recording the sensor data directly. This could then be used to predict errors in the system \cite{nentwich2020data}. In the paper, they mount vibration sensors on robots to collect data on faulty motors. Different machine learning algorithms are then tested. They are able to correctly classify faulty motors based on the data. In our current system, we do not record sensor data from the robot, but this could be added in future work. This could potentially be used to classify the success of a grasp without using vision.

Another focus is the handling of collected data. CORE  \cite{beksi2015core} is a recognition engine, where the relevant data is processed in a cloud server. This offloads the computational requirements from the workcell, and data storage, processing, and training are performed in the cloud. We currently perform all processing locally to simplify the set-up. In future work, data storage and network fine-tuning could be performed in the cloud. This could reduce the computational requirements for the robotic set-up, and allow for easier data sharing in a multi-robot set-up.

Another method for handling data is RoboFlow \cite{lin2022roboflow}. The architecture of RoboFlow is a central data engine communicating with containerized modules. The modules consist of tasks such as data preprocessing and algorithm development. The containerization makes data reuse and creating new tasks much simpler. Our current focus is on a single workcell performing the task of bin-picking, with different objects. We, therefore, do not containerize the tasks of the system. If our developed system should be diversified to perform different tasks a structure such as RoboFlow could be beneficial.
%
%
A separate task is to determine how to best train with the collected data. 
One approach is presented with the Bridge dataset \cite{ebert2021bridge}. The paper does not propose a method for data collection, but instead, a methodology to improve performance when only a small amount of training data is available. The Bridge dataset is a large dataset that is included in the training when learning a specific action. The paper demonstrates that performance for the specific task increases when the large dataset is included during training. We use the same technique when fine-tuning the network by including the large dataset from KeyMatchNet \cite{hagelskjaer2023keymatchnet}. We also test the methodology by including data from the other objects when fine-tuning, and demonstrate the effectiveness of this approach.


%


\section{Method}
\label{met}


The developed method is a data engine for self-supervised learning of pose estimation, implemented on a robotics workcell. In the following section, a full description of the developed method is given. Starting with a description of the workcell, and then the data engine is elaborated.

\subsection{Workcell} \label{met:work}

The workcell is based on the system presented in \cite{hagelskjaer2024off}. The system was tested on a replication of the bin-picking challenge presented at the 2018 World Robot Summit \cite{yokokohji2019assembly} with state-of-the-art results. One of the aspects that allowed the system to obtain good performance is the robustness of the grasping. The system is robust to failed grasps, and can simply retry until a successful grasp is obtained. This allows the system to work, even when some of the pose estimations fail, which is essential for our self-supervised method. The complete workcell is shown in Fig.~\ref{fig:workcell}.


The task is the insertion of objects, and thus the position of the objects needs to be very precise. However, if the pose estimation is incorrect or if the object is moved during the grasping, the resulting grasp pose will be erroneous. An in-hand pose estimation system has, therefore, been added to obtain precise pose estimations after grasping. The in-hand pose estimation system is elaborated in subsection~\ref{sec:inhand}. 

\begin{figure}[t]
    \vspace{1.5mm}
    \begin{center}
       \includegraphics[trim=0 70 0 270, clip,width=0.85\linewidth]{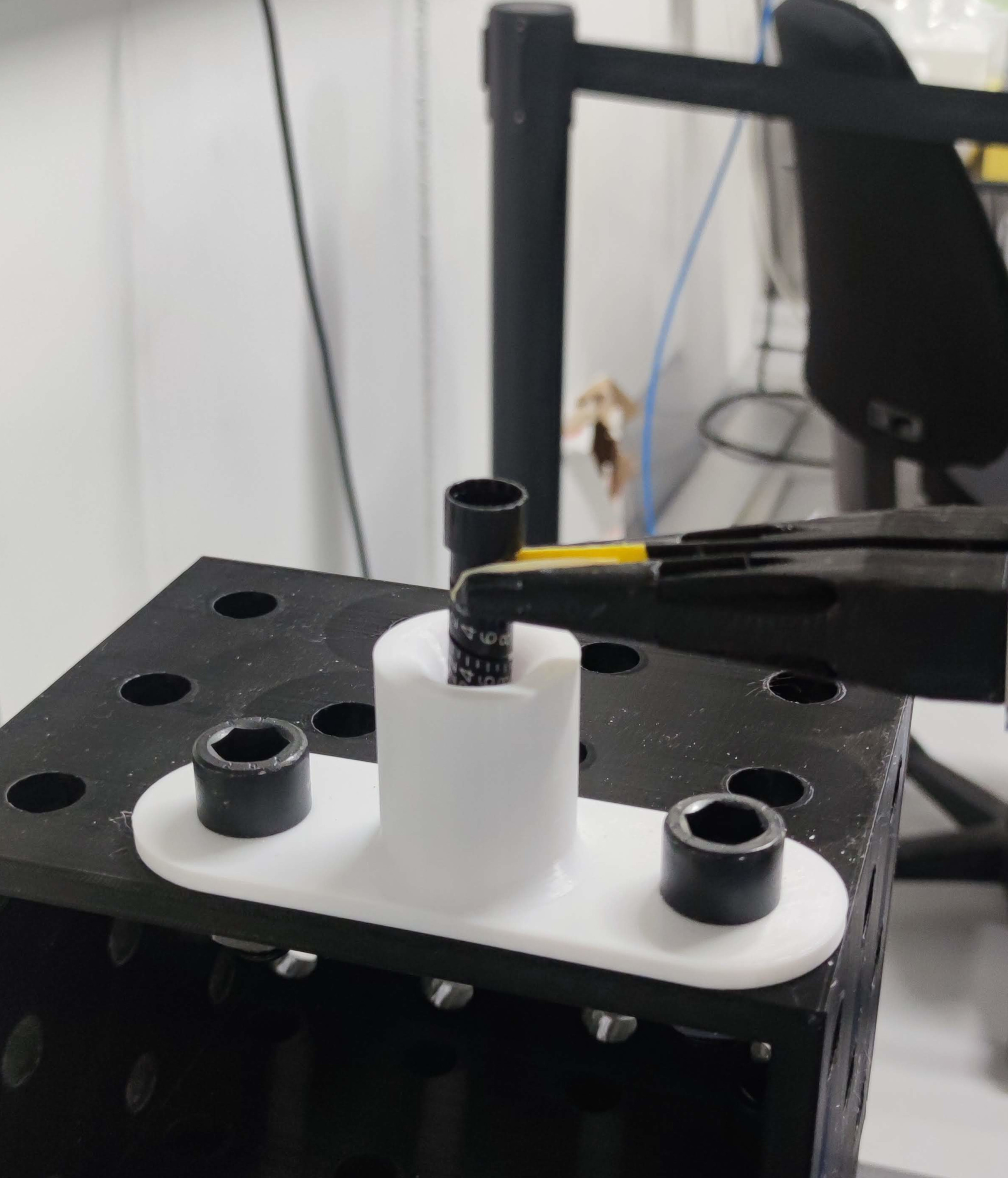}
       \caption{Insertion of Novo A into the fixture. }
       \label{fig:insertion}
        \vspace{-6mm}
    \end{center}
\end{figure}


Additionally, fixtures for the delivery of objects have been added to the workcell, these allow for a final delivery of the objects. An example of an insertion into the fixture is shown in Fig.~\ref{fig:insertion}. 
%


\subsubsection{Pose Estimation}

The pose estimation is based on an adapted version of ParaPose \cite{hagelskjaer2022parapose} used in the original workcell \cite{hagelskjaer2024off}. The pose estimation method uses only depth information as the CAD models seldom have color information. 
The ParaPose method employs a DGCNN network \cite{dgcnn} to compute the pose estimation, but it does not only perform a single pose estimation. Instead, multiple pose estimations are performed. A depth check is then used to remove wrong poses and a non-maximum suppression is used to remove duplicates. Thus even if several pose estimations fail, a successful pose can still be obtained. By increasing the number of pose estimations the possibility of a successful pose is increased, however, the run-time of the algorithm is also increased. When running the workcell the number of pose estimations is set to 48 to increase the possibility of obtaining data. As the self-supervised method increases the performance the number of pose estimations could be reduced, while retaining the performance.

The DGCNN network \cite{dgcnn} is, however, replaced with KeyMatchNet \cite{hagelskjaer2023keymatchnet}. KeyMatchNet is a zero-shot network and thus allows the system to operate immediately without a need for training and generating training data. Initially, the trained model presented in \cite{hagelskjaer2023keymatchnet} is used, however, as training data is obtained the model is fine-tuned. 



\subsubsection{Expected Grasp Pose}
The grasp poses are created as in the original system \cite{hagelskjaer2024off}. By combining the grasp poses with the pose estimations, a huge number of possible grasp solutions are created. By using the robot collision model a large number of these solutions are rejected and only feasible solutions remain \cite{hagelskjaer2024off}. 
The system then selects the shortest solution in joint space, and a grasping is attempted. This grasping approach is going for the grasp most likely to succeed, a pure exploitation strategy. While this could be changed to give a more diverse set of grasps, it is prioritized that the system runs as effectively as possible from the beginning.
As the grasping solution is created by combining a pose estimation with a grasp pose, the expected grasp pose of the object is known. This expected grasp pose is used further in the created data engine.


%

\subsubsection{In-hand Pose Estimation}
\label{sec:inhand} 

\begin{figure}[t]
    \vspace{1.5mm}
    \begin{center}
    \begin{subfigure}[t]{.23\textwidth}
      \centering
      \includegraphics[width=0.99\linewidth]{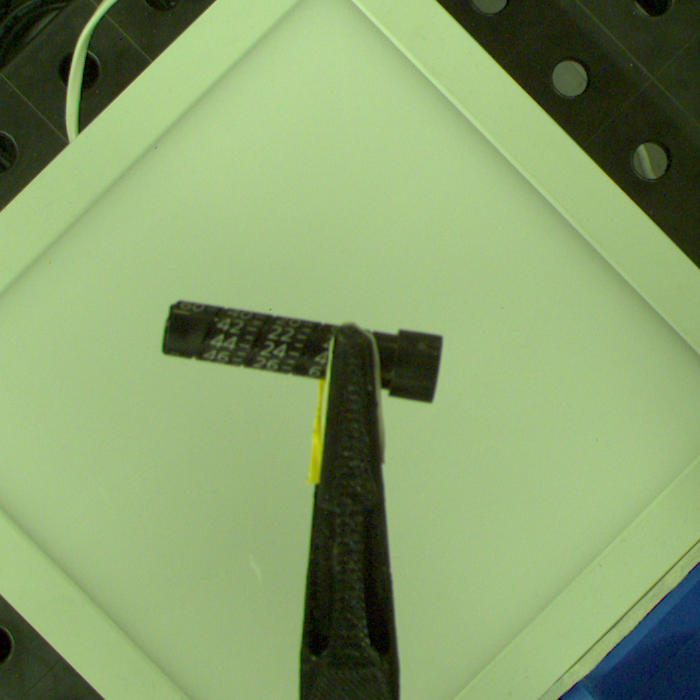}
      \caption{Correct grasp.}
      \label{fig:graspinhand:01}
    \end{subfigure}%
    ~
    \begin{subfigure}[t]{.23\textwidth}
      \centering
      \includegraphics[width=0.99\linewidth]{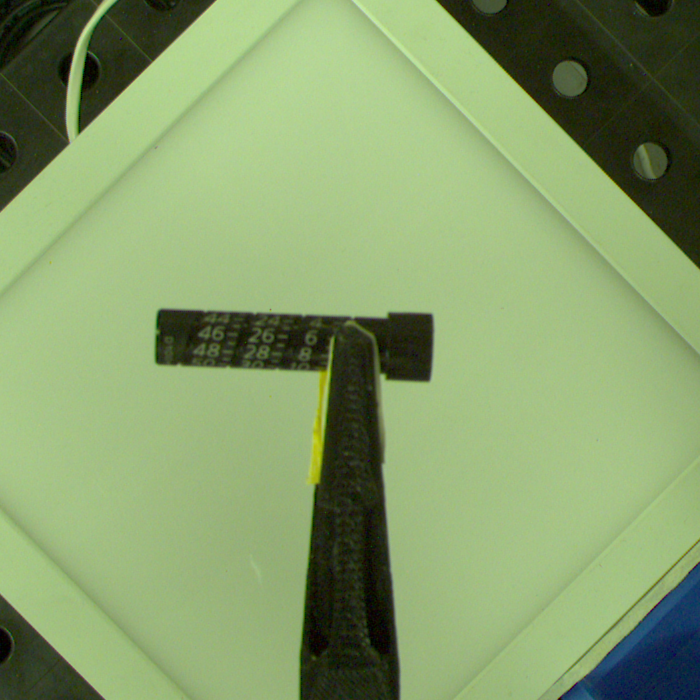}
      \caption{Minor angular error.}
      \label{fig:graspinhand:02}
    \end{subfigure}%

    ~~~
    
    ~

    \begin{subfigure}[t]{.23\textwidth}
      \centering
      \includegraphics[clip,width=0.99\linewidth]{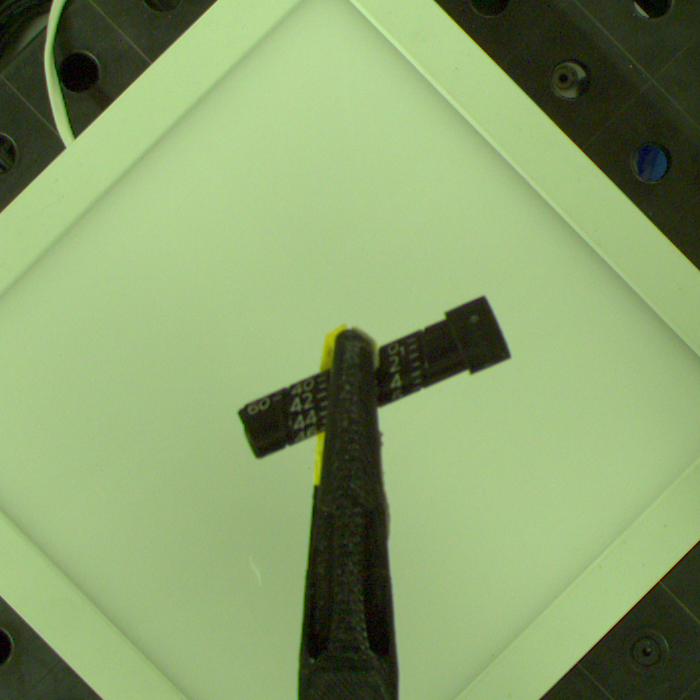}
      \caption{Large angular error.}
      \label{fig:graspinhand:04}
    \end{subfigure}%
    ~
    \begin{subfigure}[t]{.23\textwidth}
      \centering
      \includegraphics[width=0.99\linewidth]{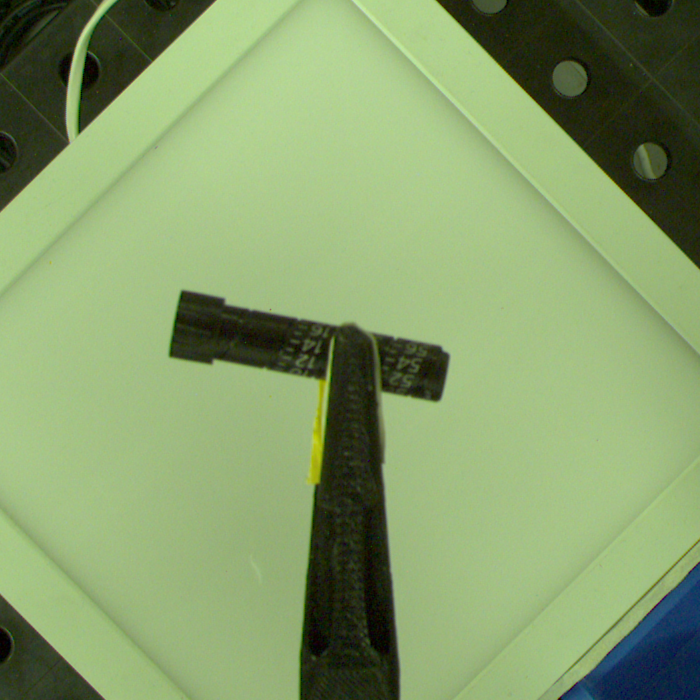}
      \caption{Wrong orientation.}
      \label{fig:graspinhand:03}
    \end{subfigure}%
    
   \caption{Examples of different grasps as shown from the in-hand vision system. Image a) and b) where successfully inserted, while c) and d) could not. }
   \label{fig:graspinhand}
    \vspace{-6mm}
\end{center}
\end{figure}

The in-hand pose estimation is using an existing method \cite{hagelskjaer2022hand}. This method combines classic template matching with stable poses. The stable pose results from the grasp which limits the possible poses of the object. The object is thus reduced to translation in x and y, and rotation about z, with respect to the camera. The cylindrical shape of the objects allows them to also rotate about their own z-axis while in the fingers. However, this rotation does not impact the insertion or the visual representation of the image. Examples of in-hand inspection after different grasps of Novo A are shown in Fig.~\ref{fig:graspinhand}.

When performing the in-hand pose estimation, the fingers grasping the object are placed 60~cm from the overhead camera, this is a known position $_{cam}T^{tcp}$. This known position is combined with the expected grasp pose of the object, $\ _{tcp}T^{obj}_{inhand}$, to create the pose used to generate the templates $_{cam}T^{obj}_{inhand}$. The equation is shown in Eq.~\ref{eq:inhandpose}:

\begin{equation} \label{eq:inhandpose}
 \  _{cam}T^{obj}_{inhand}  =  \ _{cam}T^{tcp} \ _{tcp}T^{obj}_{inhand} 
\end{equation}


\subsection{Self-supervised learning} \label{met:data}
%
The self-supervised learning is enabled by a data engine, consisting of two parts, data collection, and data labeling. As the workcell is running, and data is collected and labeled, the network is concurrently fine-tuned and improved. In the following section, the data collection, data labeling, and fine-tuning are elaborated.





\subsubsection{Data Collection}

\begin{figure}[t]
    \vspace{1.5mm}
    \begin{center}
       \includegraphics[width=0.8\linewidth]{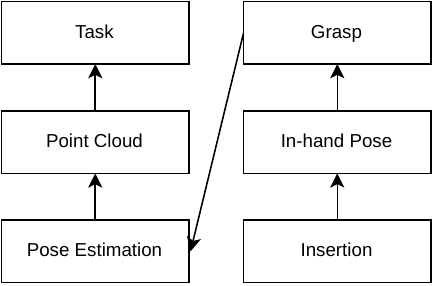}
       \caption{The layout of the database used for collecting the data. The database structure mimics the flow of the task allowing for a simple relationship between pose estimations and the resulting insertions.}
       \label{fig:database}
       \vspace{-6mm}
    \end{center}
\end{figure}

The training data for the pose estimation is a point cloud with pose annotations for the object. Instead of manually labeling the poses, we use the poses found from the pose estimation. Every time a pose estimation is performed this data is saved in a database. 
The database is created following the "tree" pattern \cite{haraty2013relational}. Meaning all the operations are automatically recorded in a parent-child relationship. The database mimics the structure of the task. Starting with the "Task" containing information about the current task, e.g. object, network type, at the top. Then all point clouds were obtained during the task, then pose estimations, grasps, in-hand pose estimations, and finally insertions. The tree structure disallows cyclical relationships simplifying look-ups. Thus an easy relationship can be made between the position of an in-hand pose estimation and the preceding object pose estimation in the bin. The database structure is shown in Fig.~\ref{fig:database}.

\subsubsection{Data Labeling}
As the system is running, it is continually obtaining pose estimations of objects in the bin. However, many of these pose estimations are expected to be incorrect. As a result the data will not only consist of true samples, but also false samples. 

\textit{Automatic Data Labeling: } While the obtained data could be sorted manually, this would severely hinder the usability of the developed method.
Instead, a auto-labeling strategy is employed using the collected data.
The auto-labeling strategy uses the in-hand pose for verification. As the in-hand pose estimation use the stable-grasp and the background-light, it is both able to eliminate clutter and occlusion, and limit the pose estimation to 3 Degrees of Freedom. As a result these poses are very precise.
%

The found in-hand pose, $_{tcp}T^{obj}_{inhand}$, is then compared with the expected grasp pose, $_{tcp}T^{obj}_{grasp}$. The expected grasp pose is the planned pose for grasping the object.
%
If an object pose is correct, the expected grasp pose should match the found in-hand pose.

The grasp pose comparison is performed using the metric from \cite{tejani2014latent}. As the objects are cylindrical the $e_{ADI}$ score is used \cite{hinterstoisser2012model, hodan2018bop}. The $e_{ADI}$ score compares the average minimum point distance between the two point clouds. Additionally, to ensure that the orientation of z-axis of the object is correct, an angular check is used. The equation is shown in Eq.~\ref{eq:addiscore2}, where $e_{\theta}$ is the angular error. The distance threshold is set to 2mm, this is much lower than in the original metric \cite{hinterstoisser2012model, hodan2018bop}. However, this is done to only accept very precise pose estimates.


\begin{equation} \label{eq:addiscore2}
 TP(e_{ADI}, e_{\theta}) = 
\begin{cases}
    1, & \text{if } e_{ADI} < 2.0mm \wedge e_{\theta} < 15 \degree \\
    0,              & \text{otherwise}
\end{cases}
\end{equation}

As many of the errors occur as a result of the object pose being rotated $180\degree$ a method to use these poses is implemented. Before calculating the overlap in Eq.~\ref{eq:addiscore2}, if the angular error is larger than $90\degree$ the object is flipped $180\degree$ about the center. This is performed both for the in bin pose and the expected grasp pose. 
The equation is shown in Eq.~\ref{eq:rotate}, where $R$ is the rotation matrix and $R_y$ is a rotation matrix $180\degree$ about the y-axis.

\begin{equation} \label{eq:rotate}
 f(R, e_{\theta}) = 
\begin{cases}
    \ R R_y , & \text{if } e_{\theta} > 90\degree \\
    \ R, & \text{otherwise}
\end{cases}
\end{equation}

\textit{Good Grasps Only: } From the automatic labeling the true and false samples are sorted into True
Positives (\textit{TP}) and True Negatives (\textit{TN}). But, any movement of objects during grasping will result in False Positives (\textit{FP}) and {FN}. 
Additionally, as only a single object in the bin is labeled using the auto labeling system all remaining objects are \textit{FN}.

But, as a result of the bin-picking scenario and the KeyMatchNet \cite{hagelskjaer2023keymatchnet} algorithm  all \textit{FN} can be completely ignored.
This is because KeyMatchNet is only trained on positive samples, as it simply performs instance segmentation and keypoint prediction. And as the task is homogeneous bin-picking, there is no need for detection or object class distinction. The position of the bins are known and the contents of the bin is known. The task is simply to estimate the pose of the objects for bin-picking.
%
%

Thus we only need to verify \textit{TP} from the training data. And all negatives, regardless of them being \textit{TN} or \textit{FN} can be discarded. The only loss occurring by discarding \textit{FN} is less training samples, and thus potentially a more biased data set. 
%

Any collision during grasping could make a perfect object pose not match the in-hand pose, resulting in a \textit{FN}, which is simply discarded. On the other hand \textit{FP} are very unlikely to occur. They only happen; if by a collision a wrongly pose estimated object is turned into a perfect grasp. The extremely rare occurrence of \textit{FP} makes their contribution negligible. 
%
%
%
The usage of data instances according to their predicted condition is shown in Tab.~\ref{tab:tfpn_table}.



\begin{table}[t]
    \vspace{1.5mm}
        \caption{Confusion matrix showing the usage of different instances according to their condition.}
        \vspace{-1.5mm}
        \hspace*{-0.25cm}
        \begin{tabular}{llcc}
                                                     &                               & \multicolumn{2}{c}{Correct pose estimation}                                      \\ \cline{3-4} 
                                                     & \multicolumn{1}{l|}{}         & \multicolumn{1}{c|}{True}                   & \multicolumn{1}{c|}{False}         \\ \cline{2-4} 
        \multicolumn{1}{l|}{In-hand pose estimation} & \multicolumn{1}{l|}{Positive} & \multicolumn{1}{c|}{\textbf{Training Data}} & \multicolumn{1}{c|}{Very Unlikely} \\ \cline{2-4} 
        \multicolumn{1}{l|}{matching expected grasp} & \multicolumn{1}{l|}{Negative} & \multicolumn{1}{c|}{Discarded}              & \multicolumn{1}{c|}{Discarded}     \\ \cline{2-4} 
        \end{tabular}
    \label{tab:tfpn_table}
    \vspace{-5mm}
\end{table}



    

\subsubsection{Network Training} \label{met:grasp}
When training data has been obtained the network is fine-tuned using the original KeyMatchNet network as a starting point. The same hyper-parameters used for training the original network are also used for the fine-tuning. 

The general approach for fine-tuning is to freeze the network, except for the last layers which are randomized, and then retrained \cite{weiss2016survey}. However, to retain the zero-shot abilities of the network this method is not applicable. Additionally, as the original network was trained on synthetic data, by including the newly collected real data the features could incorporate information from the scenario. This would allow the method to perform better on new objects in the same workcell. The network is, therefore, trained without freezing the weights, and to avoid over-fitting to the new smaller dataset, the original dataset is included during training \cite{ebert2021bridge}. For a fair comparison of the number of training samples obtained from the data engine, each epoch always consists of 2000 samples from both datasets.

As only a few samples of the new object are present, the key-point sampling is randomized during training. For $40\%$ of the training samples, random key-points are sampled instead of using the Farthest point sampling. This allows the key-point positions to be spread more randomly on the object, making the distinction more difficult for the network.

\section{Experiments}
\label{exp}

To test the effectiveness of the developed data-engine several different experiments have been performed.

\subsection{Test Objects}

\begin{figure}[t]
    \vspace{1.5mm}
    \begin{center}
       \includegraphics[width=0.8\linewidth]{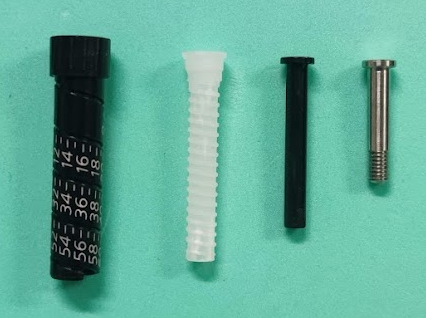}
       \caption{The four objects used for experiments. From left to right: Novo A, Novo B, Novo C and WRS Screw. The longest object is 61mm and the shortest is 32mm.}
       \label{fig:testobjects}
        \vspace{-6mm}
    \end{center}
\end{figure}

The test objects are four different cylindrical objects, three Novo Nordisk components, and a screw from the WRS 2018 Assembly Challenge \cite{yokokohji2019assembly}. The objects are shown in Fig.~\ref{fig:testobjects}. The surface of the Novo Nordisk objects is plastic, while their color differ. One black with white texture, one white and one black, and the WRS screws surface is shiny metallic. These different surface properties are expected to result in unique artifacts in the resulting point clouds. This along with the different sizes of the objects will result in varying results for the pose estimation algorithm.

The objects are all cylindrically shaped, an object type that is very prevalent in the industry. One important task with such objects is the grasping and insertion from unknown poses \cite{yokokohji2019assembly}. For the insertion to be successful the position should be very precise. 
While small errors in the pose can be compensated by the in-hand pose estimation system, many grasps poses do not allow insertion of the object. If e.g. the object is grasped too close to the bottom or if the object is rotated it cannot be inserted. A good initial pose estimation is, therefore, necessary.

\subsection{Data Collection}
In a real application, the data would be collected as the system is executing a task. However, for this paper, a data collection was performed to obtain the data. Data was collected for each of the four objects by running the bin-picking and in-hand inspection until 1000 training samples and 200 test samples had been obtained.  

During the data collection, the objects are placed in two bins, one with thirty objects and another with ten objects. Starting with the filled bin twenty objects are grasped and moved to the bin with ten objects. This continues until all the samples have been obtained or if the bins become empty. If the bins become empty they are refilled, at the start of the experiment and the data collection continues. The bin can become empty for two reasons, one reason is if the objects are dropped after grasping, as a result of an unstable grasp. Another reason is if the robot pushes objects out of the bin while grasping.

\subsection{Pose Estimation Performance}
To demonstrate the effectiveness of the developed method, the pose estimation performance is shown. Performance is shown as a result of the number of training samples used for fine-tuning. Results are shown for zero-shot pose estimation and using 1, 10, 20, 500, and 1000 training samples for fine-tuning. For comparison, we also show results for the ParaPose network, trained on synthetic data \cite{hagelskjaer2024off}.

The pose estimation is tested using the same method as in KeyMatchNet \cite{hagelskjaer2023keymatchnet}. Here a single point cloud is processed by the network, and using Kabsch-RANSAC \cite{kabsch1976solution,fischler1981random} the pose estimation is found.  
As the test data was obtained using the full ParaPose \cite{hagelskjaer2022parapose} method with multiple pose estimations and depth checks, the single pose estimation used in testing is not expected to pose estimate the object perfectly. 



\begin{figure}[t]
    \vspace{2.5mm}
    \centering
    \begin{tikzpicture}
    \begin{axis}[%
        xtick={0.05,0.19,1,10,20,200,500,1000},
        xticklabels={Synth,0,1,10,,200,,1000},
        xmode=log,
        log ticks with fixed point,
        ylabel=\textbf{Pose Estimation Recall},
        scatter/classes={%
            a={mark=o,draw=black, mark size=3.0pt}, 
            b={mark=x,draw=black, mark size=5.0pt}, 
            e={mark=diamond,draw=black, mark size=5.0pt}, 
            c={mark=triangle,draw=black, mark size=5.0pt}, 
            f={mark=*,draw=black, mark size=2.0pt}, 
            d={mark=square,draw=black, mark size=3.0pt} 
            },
            legend columns=3,
            legend entries={
            Novo A, Novo B,
            Novo C,
            WRS Screw,
            Zero-shot,
            ParaPose%
        },
        legend to name=myfancyname,xlabel={\textbf{Training Samples}}]
        legend pos=south east,]
            
    \addplot[scatter, 
        scatter src=explicit symbolic]%
        table[meta=label] {
            x y label
            0.05 0.765  d
            0.19 0.6145 f
            0.19 0.6145 a
            1 0.7745 a
            10 0.8215 a
            20 0.829 a
            200 0.936 a
            500 0.9675 a
            1000 0.9555 a
        };

    \addplot[scatter,
        scatter src=explicit symbolic]%
        table[meta=label] {
            x y label
            0.05 0.585 d
            0.19 0.377 f
            0.19 0.377 b
            1 0.3665 b
            10 0.3115 b
            20 0.313 b
            200 0.6215 b
            500 0.7105 b
            1000 0.767 b
        };

    \addplot[scatter,
        scatter src=explicit symbolic]%
        table[meta=label] {
            x y label
            0.05 0.18 d
            0.19 0.243 f
            0.19 0.243 c
            1 0.3005 c
            10 0.469 c
            20 0.5205  c
            200 0.575 c
            500 0.5645   c
            1000 0.6035 c
        };
        
    \addplot[scatter,
        scatter src=explicit symbolic]%
        table[meta=label] {
            x y label
            0.05 0.575 d
            0.19 0.345 f
            0.19 0.345 e
            1 0.1965 e
            10 0.3575 e
            20 0.395 e
            200 0.6455 e
            500 0.587 e
            1000 0.6125 e
        };
    
    \end{axis}
    \node[anchor=north] at (current axis.below south) {\ref{myfancyname}};
    
    \end{tikzpicture}
    \caption{Pose estimation accuracy as a result of number of training samples, for the four objects. To the leftmost results are also shown for ParaPose \cite{hagelskjaer2022parapose} trained with synthetic data of the CAD models.
    }
    \label{fig:poseestimationperformanceall}
    \vspace{-4mm}
\end{figure}
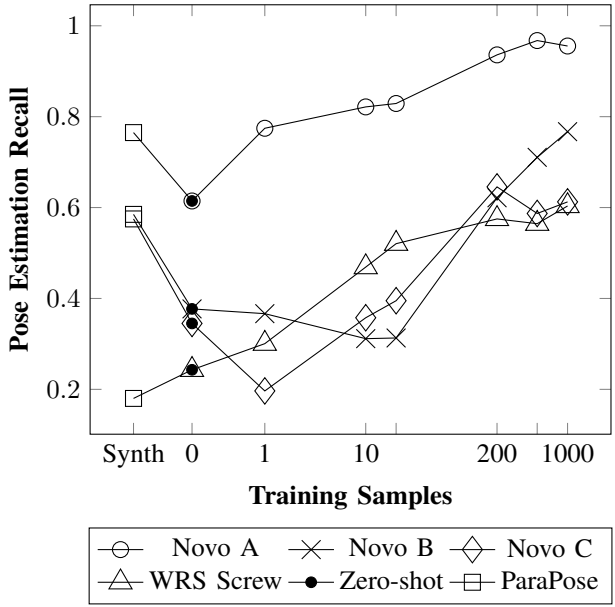

The results are shown in Fig.~\ref{fig:poseestimationperformanceall}. It is seen, that for all objects there is a correlation between the number of samples and the pose estimation recall. It is also interesting how much the performance increases with only a few amount of samples, with even a single sample showing improvement. For all objects, at 1000 samples the performance using self-supervised learning outperforms ParaPose trained on synthetic data. 
However, the performance for each object differs very much. At 1000 samples the recall for Novo A is 0.98, while Novo C is only 0.67, and WRS Screw is 0.64. But, for each the growing number of samples results in growing performance. It is thus seen that the data-engine is a viable method for a set-up to instantly start working, and then gradually increase performance.

\subsection{Cross-Validation Generalizability}
A cross-validation experiment is performed to test the method's generalizability to new objects. By learning scenario characteristics with the data engine, it is expected that the performance of novel objects will be improved. In this experiment, we show performance for each object, where the network is fine-tuned using the other objects. We also show performance for the pose estimation when the network is fine-tuned on all four objects. The result of the experiment is shown in Tab.~\ref{tab:cross}.

From the results, several interesting observations are seen. By including the other objects during training a better performance is obtained for all objects, except Novo A which shows slightly lower performance. An even more significant finding is the performance when the test object is excluded in the fine-tuning. For all objects, the performance is better compared with the non-tuned network. And, for two objects, the performance is actually better compared with training only on the test object. This indicates that the network generalizes to the scenario and not only learns object-specific features. When introducing novel objects to the workcell, the previous processing of other objects will thus help improve the performance.

\begin{table}[t]
    \vspace{1.5mm}
    \centering
        \caption{Pose estimation recall as a result of fine-tuning when 500 training samples have been obtained for each object. Results are shown without any fine-tuning, fine-tuning with the test object, fine-tuning with the test object excluded, and with all four objects for fine-tuning.}
    \begin{tabular}{|c|c|c|c|c|}
        \hline
            & Novo A & Novo B & Novo C & WRS Screw \\ \hline
         ParaPose \cite{hagelskjaer2022parapose}
                            & 0.77 & 0.59  & 0.58 & 0.18  \\ \hline
         Zero Shot          & 0.62 & 0.38 & 0.35 & 0.24 \\ \hline
         Object Excluded    & 0.66 & 0.50 & 0.62 & 0.69 \\ \hline
         Test Object Only   & 0.97 & 0.71 & 0.59 & 0.57 \\ \hline
         All Objects        & 0.95 & 0.74 & 0.66 & 0.64 \\ \hline
    \end{tabular}
    \label{tab:cross}
\end{table}




\subsection{Grasping and Insertion}
The task of the workcell is the correct insertion of the objects. To determine the data engine's ability in this regard, an insertion experiment has been performed. The experiment is performed using the object Novo A. We perform the full pipeline with pose estimation, grasping, in-hand inspection, and insertion. 
For this test, the full pose estimation method is used with multiple pose estimations and the depth check. We show results for the zero-shot method and with the fine-tuned network using 1000 samples. We also show results for the method using 48 and 6 pose estimations. 48 is the number of pose estimations used during the data collection.


%
The results are shown in Tab.~\ref{tab:wrs}. From the results, it is seen that the grasping performance is very similar for all configurations. Only around five percent of found objects are not grasped successfully. However, this does not indicate if the objects are grasped erroneously. When comparing with the success rate of the insertions, the performance varies much more. The self-supervised method vastly outperforms the zero-shot approach, even with a batch-size of only six. Thus by using the data engine the performance of the algorithm could be improved while reducing the run-time. The developed method is thus able to improve the performance of the robotic workcell. 

\begin{table}[t]
    \begin{center}
    \caption{
    The success rate of insertions for the Novo A object, as a result of different numbers of pose estimations used. Results are shown without any training data and using 1000 training samples.
    }
    \begin{tabular}{|l|c|c|c|c|}
        \hline
        \begin{tabular}{@{}c@{}}  \end{tabular} & \multicolumn{2}{c|}{\textbf{Zero-Shot}} & \multicolumn{2}{c|}{\textbf{Self-supervised}}       \\
        \hline
        Pose Estimations    & 48 & 6 & 48 & 6 \\ 
        \hline
        Grasping            & 95.9  &  95.6     & 95.5 & 95.4 \\
        \hline
        Insertion           & 73.1  &  76.0     & 86.7 & 82.0 \\ \hline
        
        
        
        \end{tabular}
        \label{tab:wrs}
    \end{center}
    \vspace{-4mm}
\end{table}

\section{Conclusion}
\label{con}

In this paper, we have presented a novel data engine for self-supervised fine-tuning of pose estimation. The data engine allows a workcell to operate without configuration while gradually improving performance. It is demonstrated that the self-supervised method outperforms a state-of-the-art method trained on the object CAD model.

Additionally, tests on four different objects have shown that the self-supervised learning generalizes to novel objects. It is thus possible for the workcell to start with better performance if it has already processed other objects.






In further work, an exploration strategy could be used when selecting objects to grasp. This could improve the variability of the training data.
Other labeling strategies could also be integrated to improve the data collection, using e.g. multi-view to verify correct poses instead of only using the grasping and in-hand pose estimation.

The developed framework can be used to gather data from many different objects. This would allow the training of a more general object pose estimation using the scene information such sensor and object characteristics.



\bibliographystyle{IEEEtran} 
\bibliography{egbib}

\end{document}